\documentclass{article}
     \PassOptionsToPackage{numbers}{natbib}

\usepackage[final]{tackling_climate_workshop_style}

\usepackage{ISG_style}

\usepackage[utf8]{inputenc} 
\usepackage[T1]{fontenc}    
\usepackage{hyperref}       
\usepackage{url}            
\usepackage{booktabs}       
\usepackage{amsfonts}       
\usepackage{nicefrac}       
\usepackage{microtype}      

\usepackage{graphicx}
\usepackage{multirow}
\usepackage{enumitem}
\usepackage[table]{xcolor}

\usepackage{algorithm}
\usepackage{makecell}
\usepackage{algcompatible}
\usepackage{algpseudocode}

\usepackage{amsmath,amssymb} 
\usepackage{dsfont}

\DeclareMathOperator*{\argmax}{arg\,max}

\newcommand{\lodf}{{\sf LODF}}

\newcommand{\maxAction}{{\sf maxAction}}
\newcommand{\maxReward}{{\sf maxReward}}

\newcommand{\topFiveActions}{{\sf topFiveActions}}
\newcommand{\noop}{{\sf Do\text{-}Nothing}}
\newcommand{\reco}{{\sf Re\text{-}Connection}}

\title{RL for Mitigating Cascading Failures: Targeted Exploration via Sensitivity Factors}

\author{%
    Anmol~Dwivedi\\
    Rensselaer Polytechnic Institute
    \And
    Ali~Tajer \\
    Rensselaer Polytechnic Institute
    \And
    Santiago Paternain\\
    Rensselaer Polytechnic Institute
    \And
    Nurali~Virani\\
    GE Vernova Advanced Research
}

\begin{document}

\maketitle

\vspace{-0.25in}

\begin{abstract}
Electricity grid's resiliency and climate change strongly impact one another due to an array of technical and policy-related decisions that impact both. This paper introduces a physics-informed machine learning-based framework to enhance grid's resiliency. Specifically, when encountering disruptive events, this paper designs remedial control actions to prevent blackouts. The proposed~\textbf{P}hysics-\textbf{G}uided \textbf{R}einforcement \textbf{L}earning (PG-RL) framework determines effective real-time remedial line-switching actions, considering their impact on power balance, system security, and grid reliability. To identify an effective blackout mitigation policy, PG-RL leverages power-flow sensitivity factors to guide the RL exploration during agent training. Comprehensive evaluations using the Grid2Op platform demonstrate that incorporating physical signals into RL significantly improves resource utilization within electric grids and achieves better blackout mitigation policies -- both of which are critical in addressing climate change.
\end{abstract}

\vspace{-0.2in}

\section{Introduction}
\label{sec:Introduction}

Power grid resiliency and climate change are symbiotically interconnected. Climate change is increasing the frequency and intensity of extreme weather events, such as hurricanes, floods, wildfires, and heatwaves, requiring improved grid resiliency to maintain power and reduce economic and societal impacts. Mitigating climate change needs reduction in the energy system's carbon footprint, which critically hinges on integrating renewable resources at scale. However, grid resilience enhancement is needed to provide robustness against equipment failures and manage stability impact of variability from renewable generation. Thus, mitigating and adapting to climate change necessitates enhancing grid resilience. This paper provides a physics-informed machine learning (ML) approach to enhance grid resiliency, defined as the grid's ability to withstand, adapt, and recover from disruptions.

One major source of disruption impacting grid resiliency are transmission line and equipment failures, often caused due to aging infrastructure stressed by extreme weather and congestion due to growing electricity demand. These gradual stresses can lead to system anomalies that can escalate if left unaddressed~\cite{NAPS:2004}. To mitigate these risks, system operators implement real-time remedial actions like network topology changes~\cite{Fisher:2008, Khodaei:2010, Fuller:2012, Dehghanian:2015}. Selecting these remedial actions must balance two opposing impacts: greedy actions render~\emph{quick} impact to protect specific components but may have inadvertent consequences, while look-ahead strategies enhance network robustness but have delayed impact. Striking this balance is crucial for maintaining reliable operation and maximizing grid utilization.

There are two main approaches for the sequential design of real-time remedial decisions: model-based and data-driven. Model-based methods, like model predictive control (MPC),~\emph{approximate} the system model and use multi-horizon optimization to predict future states and make decisions~\cite{larsson:2002, Carneiro:2010, almassalkhi:1-2014, almassalkhi:2-2014}. While these methods offer precise control by adhering to system constraints, they require an accurate analytical model, which can be difficult for T-grids. Moreover, coordinating discrete actions like line-switching over extended planning horizons is computationally intensive and time-consuming. Conversely, data-driven approaches like deep reinforcement learning (RL) learn decision policies through sequential interactions with the system model. Deep RL has been successfully applied to various power system challenges~\cite{Ernst:2004, Yan:2017, Duan:2020, GRNN:2024}. By shifting the computational burden to the offline training phase, these methods allow for rapid decision-making during real-time operations, making them promising for real-time network overload management~\cite{kelly:2020, marot:2020, marot:2021}.

Using off-the-shelf RL algorithms (method-driven algorithms~\cite{rolnick:2024}) for complex tasks like power-grid overload management presents computational challenges, primarily due to the systems' scale and complexity. Generic exploration policies often select actions that cause severe overloads and blackouts, preempting a comprehensive exploration of the Markov decision process (MDP) state space. This limitation hampers accurate decision utility predictions for the unexplored MDP states, rendering a highly sub-optimal remedial control policy. A solution to circumvent the computational complexity and tractability is leveraging the physics knowledge of the system and incorporating it into RL exploration design.

\noindent{\bf Contribution:} We formalize a~\textbf{P}hysics-\textbf{G}uided~\textbf{R}einforcement~\textbf{L}earning (PG-RL) framework for real-time decisions to alleviate transmission line overloads over long operation planning horizons. The framework's key feature is its efficient~{physics-guided} exploration policy design that judiciously exploits the underlying structure of the MDP state and action spaces to facilitate the integration of auxiliary domain knowledge, such as power-flow sensitivity factors~\cite{wood:2013}, for a physics-guided exploration during agent training. Extensive evaluations on Grid2Op~\cite{donnot:2020} demonstrate the superior performance of our framework over counterpart black-box RL algorithms. The data and code required to reproduce our results is publicly~\href{https://github.com/anmold-07/Physics-Guided-Blackout-Mitigation/tree/main/LineSwitchAgent_LODF}{available}.

\noindent{\bf Related Work:} The study in~\cite{Lan:2020} uses guided exploration based on $Q$-values while~\cite{AAAI:2023} employs policy gradient methods, both on bus-split actions pre-selected via exhaustive search. To accommodate the exponentially many bus-split topological actions, the study in~\cite{ICLR:2021} employs graph neural networks combined with hierarchical RL~\cite{sutton:1999} to structure agent training. Recent approaches, such as~\cite{Matavalam:2023} and~\cite{meppelink:2023}, focus on integrating domain knowledge via curriculum learning and combining it with Monte-Carlo tree search for improved action selection. However, existing RL approaches (i)~focus exclusively on bus-splitting actions; (ii) lack the integration of physical power system signals for guided exploration; and (iii) overlook active line-switching, particularly line~\emph{removal} actions, due to concerns about reducing power transfer capabilities and increasing cascading failure risk.

\vspace{-0.2in}

\section{Problem Formulation}
\label{sec: Problem Formulation}

Transmission grids are vulnerable to stress by adverse internal and external conditions, e.g., line thermal limit violations due to excessive heat and line loading. Without timely remedial actions, this stress can lead to cascading failures resulting in blackouts. To mitigate these risks, our objective is to maximize the system's survival time over a horizon $T$, denoted by ST$(T)$, defined as the time until a blackout occurs~\cite{donnot:2020}. In this paper, we focus on line-switching actions $\bW_{\sf line}[n] \dff [ W_1[n], \dots, W_L[n]]^{\top}$ to reduce system stress by controlling line flows, where the binary decision variable $W_{\ell}[n] \in \{0, 1\}$ indicates whether line $\ell$ is removed (0) or reconnected (1) at time $n\in[T]$. We also define $c^{\sf line}_{\ell}$ as the cost of line-switching for line $\ell$. Hence, the system-wide cost incurred due to line-switching over a horizon~$T$ is $C_{\sf line}(T) \dff \sum_{n=1}^{T} \sum_{\ell=1}^{L} c^{\sf line}_{\ell}\cdot W_{\ell}[n]$.

\noindent \textbf{Operational Constraints:} Line-switching decisions are constrained by operational requirements to maintain system security. Once a line is switched, it must remain offline for a mandated downtime period $\tau_{\rm D}$ before being eligible for another switch. For naturally failed lines (e.g., due to prolonged overload), a longer downtime period $\tau_{\sf F}$ is required before reconnection, where $\tau_{\sf F} \gg \tau_{\rm D}$.

\noindent \textbf{Maximizing Survival Time:} Our objective is to constantly monitor the system and, upon detecting mounting stress (e.g., imminent overflows), initiate flow control decisions (line-switching) to maximize the system's ST$(T)$. Such decisions are highly constrained with decision costs $C_{\sf line}(T)$ and operational constraints due to downtime periods $\tau_{\rm N}$ and $\tau_{\rm F}$. To quantify ST$(T)$, we use a proxy, the~\textbf{risk margin} for each transmission line $\ell$ at time $n$, defined as $\rho_{\ell}[n] \dff \frac{A_{\ell}[n]}{A_{\ell}^{\sf max}}$, where $A_{\ell}[n]$ and $A_{\ell}^{\sf max}$ denotes the present and maximum line current flows, respectively. Based on $\rho_{\ell}[n]$, a line $\ell$ is considered~\emph{overloaded}, if $\rho_{\ell}[n] \geq 1$. Minimizing these risk margins reduces the likelihood of overloads, thereby extending ST$(T)$. We also use risk margins to identify~\emph{critical} states, which are states that necessitates remedial interventions, defined by the rule $\max_{i \in [L]} \rho_{i}[n] \geq \eta$. To maximize ST$(T)$, our goal is to sequentially form the decisions $\bar\bW_{\rm line}\triangleq \{\bW_{\sf line}[n]:n\in\N\}$ all while adhering to operational constraints and controlled decision costs $\beta_{\rm line}$, formulated as:
\begin{equation}
    \label{eq:OBJ2}
    \mcP: \left\{
    \begin{array}{cl}
        \displaystyle 
        \min_{ \{\bar\bW_{\rm line}\} }
        & \displaystyle \sum_{n=1}^{T}  \sum_{\ell=1}^{L} \rho_{\ell}[n]\\
        {\rm s.t.}
         & C_{\rm line}(T) \leq \beta_{\rm line}\\  
         & \mbox{\footnotesize Operational Constraints}
    \end{array}\right. \ .
\end{equation}

\noindent \textbf{Cascading Failure Mitigation as an MDP:} The complexity of identifying optimal line-switching (discrete) decisions grows exponentially with the number of lines $L$ and the target horizon $T$, and is further compounded by the need to meet operational constraints. To address the challenges of solving $\mcP$ in~\eqref{eq:OBJ2}, we design an agent-based approach. At any instance $n\in[T]$, the agent has access to the system's states $\{\bX[m]:m\in[n]\}$ and uses this information to determine the line-switching actions. These actions lead to outcomes that are partly deterministic, reflecting the direct impact on the system state, and partly stochastic, representing the randomness of future electricity demands. To effectively model these stochastic interactions, we employ a Markov decision process (MDP) characterized by the tuple $(\mcS, \mcA_{\sf line}, \P, \mcR, \gamma)$. Detailed information about the MDP modeling techniques employed is provided in Appendix~\ref{sec:MDP Modeling}. Finding an~\emph{optimal} decision policy $\pi^{*}$ can be found by solving~\cite{bellman:1957} 
\begin{equation}
    \label{eq: optimal policy}
    \mcP_{2}: \quad \pi^{*}(\bS) \dff \argmax_{\pi} \; Q_{\pi}(\bS, \pi(\bS))\ ,
\end{equation} where $Q_{\pi}(\bS, a)$ characterizes the state-action value function.

\section{Physics-Guided RL Framework}
\label{sec: Physics-Guided RL Framework}

\noindent \textbf{Motivation:} Model-free off-policy RL algorithms~\cite{tsitsiklis:1997, sutton:2018} with function approximation~\cite{Mnih:2015} are effective in finding good policies without requiring access to the transition probability kernel $\P$ for high-dimensional MDP state spaces $\mcS$. However, the successful design of these algorithms hinges on a comprehensive exploration of the state space to accurately learn the expected decision utilities, such as $Q$-value estimates. Common approaches entail dynamically updating a behavior policy $\pi$, informed by a separate exploratory policy like $\epsilon$-greedy~\cite{sutton:2018}, illustrated in Algorithm~\ref{alg:ALGO1}. While $Q$-learning with~\emph{random} $\epsilon$-greedy exploration is effective in many domains~\cite{Mnih:2015}, it faces challenges in power-grid overload management. Random network topology exploration actions $a[n]\in\mcA_{\sf line}$ can quickly induce severe overloads and, thus, blackouts. This is because topological actions force an abrupt change in the system state $\bX[n]$ by redistributing transmission line power-flows after a network topological change, compromising risk margins $\rho_{\ell}$ and exposing the system to potential cascading failures, preventing a comprehensive exploration of $\mcS$. This results in inaccurate $Q$-value predictions for the unexplored MDP states, rendering a highly sub-optimal remedial control policy.

\setlength{\textfloatsep}{0.001pt}
\begin{minipage}{0.46\textwidth}
\begin{algorithm}[H]
\small
\caption{Canonical $\epsilon$-greedy Exploration}
\label{alg:ALGO1}
\begin{algorithmic}[1]
\State \textbf{Input:} $\epsilon_{1}$, $\mcA$, $Q(s, a)$, \quad \textbf{Output:} Action $a$
\If{$\mu \sim \mathcal{U}(0, 1) < \epsilon_{1}$}
    \State $a \sim \text{Uniform}(\mathcal{A})$~\Comment{\textcolor{blue}{Random-Explore}}
\Else~\Comment{\textcolor{blue}{$Q$-guided Exploit}}
    \State Select $a$ based on $Q(s, a')$
\EndIf
\end{algorithmic}
\end{algorithm}
\vspace{.38 in}
\end{minipage}
\hfill
\begin{minipage}{0.54\textwidth}
\begin{algorithm}[H]
\small
\caption{Physics-Guided $\epsilon$-greedy Exploration}
\label{alg:ALGO2}
\begin{algorithmic}[1]
\State \textbf{Input:} $\epsilon_{1}, \epsilon_{2}, \mcA, Q(s, a)$ \quad \quad \textbf{Output:} Action $a$
\If{$\mu \sim \mathcal{U}(0, 1) < \epsilon_{1}$}
    \If{$\zeta \sim \mathcal{U}(0, 1) < \epsilon_{2}$}~\Comment{\textcolor{blue}{Physics-Explore}}
        \State $a \sim \text{Physics-Guided}(\mathcal{A})$~\Comment{\textcolor{red}{Algorithm~\ref{alg:algoExplore}}}   
    \Else
        \State $a \sim \text{Uniform}(\mathcal{A})$~\Comment{\textcolor{blue}{Random-Explore}}
    \EndIf
\Else~\Comment{\textcolor{blue}{$Q$-guided Exploit}}
    \State Select $a$ based on $Q(s, a')$
\EndIf
\end{algorithmic}
\end{algorithm}
\end{minipage}

\noindent \textbf{Sensitivity Factors:} We leverage power-flow~\emph{sensitivity factors} to guide exploration decisions by augmenting $\epsilon$-greedy during agent training, as illustrated in Algorithm~\ref{alg:ALGO2}. Sensitivity factors~\cite{wood:2013} help express the mapping between MDP states $\mcS$ and actions $\mcA$ by linearizing the system around the current operating point. This approach allows us to analytically approximate the impact of any action $a[n] \in \mcA$ on risk margins and, consequently, the MDP reward $r\in \mcR$. To address the challenges associated with implementing random topological actions during $\epsilon$-greedy exploration, we use line outage distribution factors (LODF) to analyze the effects of line removals. Specifically, the sensitivity factor matrix $\lodf\in\R^{L \times L}$, represents the impact of removing line $k$ on the flow in line $\ell$ by~\cite{wood:2013}
\begin{equation}
    \label{eq: lodf}
    F_{\ell}[n+1] \approx F_{\ell}[n] + \lodf_{\ell, k}[n] \cdot F_{k}[n]\ ,
\end{equation} where $F_{k}[n]$ is the pre-outage flow in line $k$, helping predict the anticipated impact of line removal action $k$. Likewise, the sensitivities of line flows to line reconnection actions are derived in~\cite{Sauer:2001}.

\noindent \textbf{Physics-Guided Exploration:} We leverage sensitivity factors to guide agent exploration with the following key idea: Topological actions $a[n]\in\mcA_{\sf line}$ that~\textbf{reduce} line flows below their limits $A_{\ell}^{\sf max}$, without causing overloads in~\textbf{other healthy} lines, help transition to more favorable MDP states in the short term, that may otherwise be challenging to reach by taking a sequence of random exploratory actions. However, removing a line $k$ can both reduce flow in some lines and increase flow in others. To address this, we focus on identifying remedial actions that~\textbf{minimize} flow in the~\textbf{maximally} loaded line. At time $n$, we define the maximally loaded line index $\ell_{\sf max}\dff \argmax_{\ell \in [L]} \rho_{\ell}[n]$. By leveraging the structure of the $\lodf[n]$ matrix, we first design Algorithm~\ref{alg:algoLineRemoval} to identify an effective set $\mcR^{\sf eff}_{\sf line}[n]$ of potential remedial actions $a[n] \in \mcA_{\sf line}$ that greedily~\textbf{reduce} risk margin $\rho_{\ell_{\sf max}}[n]$. Then, the agent selects an action $a[n] \in \mcR^{\sf eff}_{\sf line}[n]$, guided by the dynamic effective set $\mcR^{\sf eff}_{\sf line}[n]$, as outlined in Algorithm~\ref{alg:algoExplore}, for action selection during agent training (as per the PG-RL design in Algorithm~\ref{alg:ALGO2}).

\begin{table*}[t]
	\centering
	\renewcommand{\arraystretch}{1.25} 
	\scalebox{0.75}{
		{\begin{tabular}{|c|c|c|c|c|c|c|c|}
            \hline
            \thead{Action Space $(|\mcA|)$}  
            & \thead{Agent Type}
            & \thead{Avg. ST} 
            & \thead{$\%$\\ Do-nothing}
            & \thead{$\%$\\ Reconnect}
            & \thead{$\%$\\ Removals}
            & \thead{Avg. Action Diversity}\\
            \hline
            \hline
            $-$                      &$\noop$ &$4733.96$ &$100$   &$-$     &$-$     &$-$ \\
            \hline
            $\mcA_{\sf line}\;(60)$ &$\reco$ &$4743.87$ &$99.90$  &$0.10$    &$-$ &$1.093\;(1.821\%)$ \\
            \hline
            $\mcA_{\sf line}\;(119)$ &\texttt{milp\_agent}\cite{MILPAGENT:2022}  &$4062.62$ &$12.05$ &$1.70$  &$86.24$     &$6.093\;(5.12\%)$ \\
            \hline
            \multirow{2}{*}{\rotatebox[origin=c]{0}{\makecell{$\mcA_{\sf line}\;(119)$ 
            \\ \quad $\mu_{\sf line}=0$ } }}
            &\multirow{1}{*}{\rotatebox[origin=c]{0}{\makecell{ $\pi^{\sf rand}_{\btheta}(0)$ }}} 
            &$5929.03$  &$26.78$ &$5.85$ &$67.35$  &$13.406\;(11.265\%)$\\
            \cline{2-7}
            &\multirow{1}{*}{\rotatebox[origin=c]{0}{
            $\text{PG-RL}~[\pi^{\sf physics}_{\btheta}(0)$] }} 
            &$\mathbf{6657.09}$  &$1.74$ &$7.66$ &$90.59$  &$\mathbf{17.062\;(14.337\%)}$ \\
            \hline
            \multirow{2}{*}{\rotatebox[origin=c]{0}{\makecell{$\mcA_{\sf line}\;(119)$
            \\ \quad $\mu_{\sf line}=1$}}}
            &\multirow{1}{*}{\rotatebox[origin=c]{0}{\makecell{ $\pi^{\sf rand}_{\btheta}(1)$ }}} 
            &$5327.06$  &$81.51$ &$0.28$ &$18.20$  &$3.625\;(3.046\%)$\\
            \cline{2-7}
            &\multirow{1}{*}{\rotatebox[origin=c]{0}{
            $\text{PG-RL}~[\pi^{\sf physics}_{\btheta}(1)$] }} 
            &$\mathbf{6603.56}$  &$13.93$ &$7.00$ &$79.06$  &$\mathbf{17.156\;(14.416\%)}$ \\
            \hline            
            \multirow{2}{*}{\rotatebox[origin=c]{0}{\makecell{$\mcA_{\sf line}\;(119)$
            \\ \quad $\mu_{\sf line}=1.5$}}}
            &\multirow{1}{*}{\rotatebox[origin=c]{0}{\makecell{ $\pi^{\sf rand}_{\btheta}(1.5)$ }}} 
            &$4916.34$  &$92.69$ &$0.01$ &$7.28$  &$3.406\;(2.862\%)$\\
            \cline{2-7}
            &\multirow{1}{*}{\rotatebox[origin=c]{0}{
            $\text{PG-RL}~[\pi^{\sf physics}_{\btheta}(1.5)$] }} 
            &$\mathbf{6761.34}$  &$46.53$ &$6.12$ &$47.34$  &$\mathbf{15.718\;(13.208\%)}$ \\
            \hline
		\end{tabular}}
	}
	\caption{Performance on the Grid2Op 36-bus system with $\eta = 0.95$.}
	\label{tab:lineAndGen_36}
        \vspace{0.06in}
\end{table*}

\vspace{-0.1in}

\section{Experiments}
\label{sec: Experiments}

To demonstrate our framework, we use the Grid2Op 36-bus and the IEEE 118-bus power networks from Grid2Op~\cite{donnot:2020}. Detailed descriptions of the Grid2Op dataset, environment, and performance metrics are in Appendix~\ref{sec:Grid2Op Environment Details}. We train RL agents with a~{dueling} NN architecture~\cite{wang2016dueling} with~{prioritized} experience replay~\cite{Schaul:2016} and $\epsilon$-greedy exploration. Appendix~\ref{sec:Baseline Agents} provides a thorough description of the baselines. Table~\ref{tab:lineAndGen_36} compares the agent's survival time ST$(T)$, averaged across all test episodes for $T=8062$, showing increased agent sophistication as we move down the table. We denote the best policy from random $\epsilon$-greedy (Algorithm~\ref{alg:ALGO1}) as $\pi^{\sf rand}_{\btheta}(\mu_{\sf line})$ and from physics-guided $\epsilon$-greedy (Algorithm~\ref{alg:ALGO2}) by $\pi^{\sf physics}_{\btheta}(\mu_{\sf line})$. For fair comparisons, DQN$_{\btheta}$ models for each $\mu_{\sf line}$~\eqref{eq: reward} are trained independently using Algorithms~\ref{alg:ALGO1} and~\ref{alg:ALGO2} for $20$ hours, using identical hyperparameters listed in Appendix~\ref{sec:DQN Architecture and Training}. We also adopt an exponential decay schedule for $\epsilon_{1}$ while fix $\epsilon_{2}=1$ in Algorithm~\ref{alg:ALGO2}.

In Table~\ref{tab:lineAndGen_36}, we observe that policy $\pi^{\sf physics}_{\btheta}(0)$ achieves an average ST of 6,657.09, a $12.2\%$ improvement over $\pi^{\sf rand}_{\btheta}(0)$ and a $25.2\%$ increase over baselines. Notably, the physics-guided agent takes $25.05\%$ more line-switch actions than its random counterpart, successfully identifying more effective line-removal actions due to the targeted design of $\mcR^{\sf eff}_{\sf line}[n]$ during agent training. {To illustrate this effectiveness, Fig.~\ref{fig:36bus_interactions} plots the number of agent-MDP interactions as a function of agent training time for $\mu_{\sf line}=0$. We observe that the PG-RL design results in a greater number of agent-MDP interactions, indicating a more thorough exploration of the MDP state space for the same computational budget. 

The ability of $\pi^{\sf physics}_{\btheta}$ to identify more effective actions, in comparison to $\pi^{\sf rand}_{\btheta}$, is further substantiated by incrementally increasing $\mu_{\sf line}$~and observing the performance changes. As $\mu_{\sf line}$ increases, the reward $r[n]$ in~\eqref{eq: reward} becomes~\emph{less} informative about potentially effective actions due to the increasing penalties on line-switch actions, thus amplifying the importance of physics-guided exploration design. This is observed in Table~\ref{tab:lineAndGen_36} where unlike the policy $\pi^{\sf rand}_{\btheta}(\mu_{\sf line})$, the ST associated with $\pi^{\sf physics}_{\btheta}(\mu_{\sf line})$ does not degrade as $\mu_{\sf line}$ increases. It is noteworthy that despite the inherent linear approximations of sensitivity factors, confining the RL exploration to actions derived from the set $\mathcal{R}^{\sf eff}_{\sf line}[n]$ enhances state space exploration. Overall, the agent's ability to identify impactful topological actions, leading to greater action diversity, contributes to the enhanced utilization of the electrical grid while also a significant increase in ST. Similar results for the IEEE 118-bus system are provided in Appendix~\ref{sec:118}, confirming the trends observed in the Grid2Op 36-bus system.

\begin{figure}[t]
    \centering
    \includegraphics[width=0.9\linewidth]{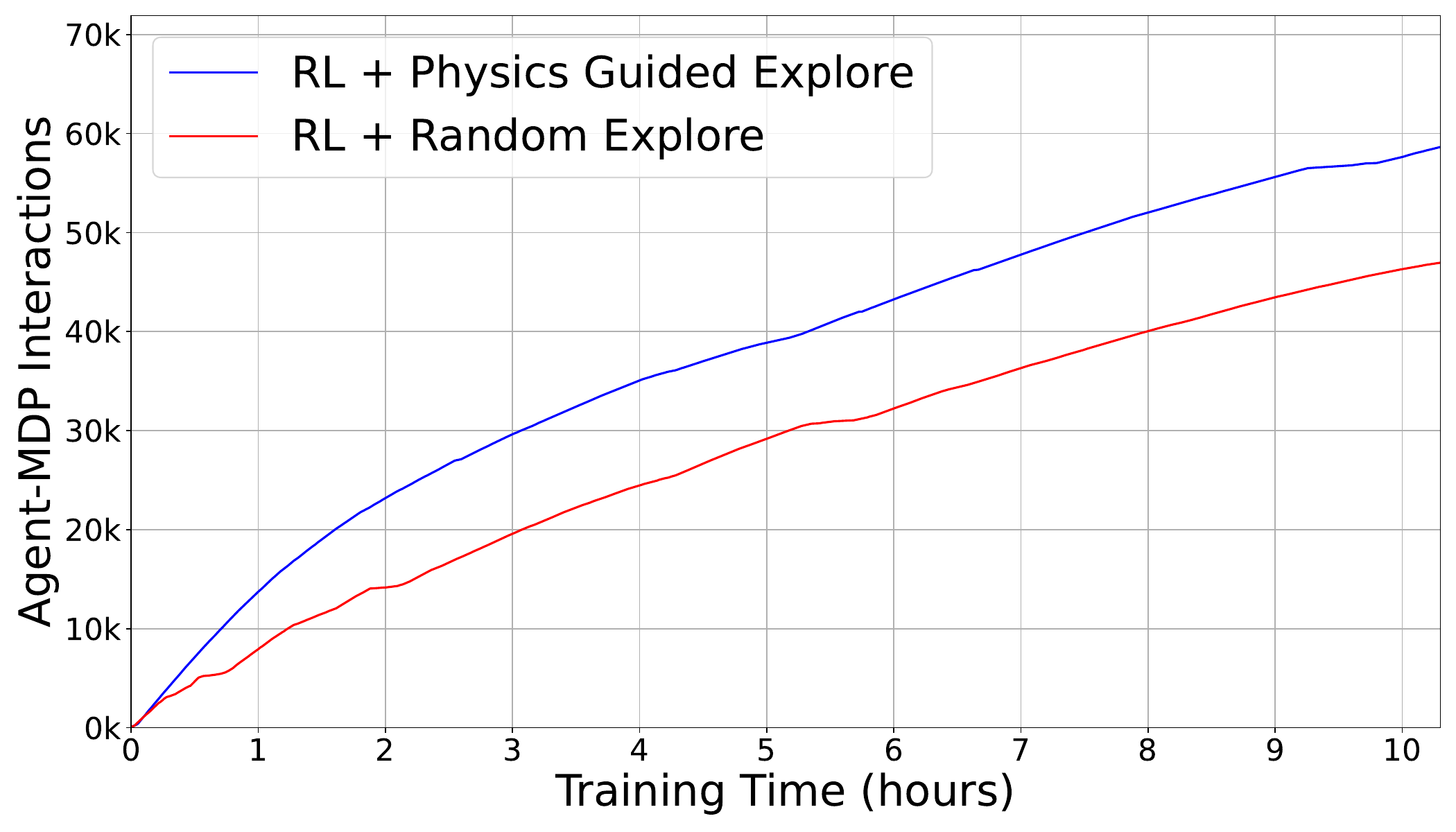}
    \caption{Agent-MDP interactions for the Grid2Op 36-bus system with $\eta=0.95$ and $\mu_{\sf line}=0$.}
    \label{fig:36bus_interactions}  
    \vspace{0.25in}
\end{figure}

\section{Conclusion and Future Work}
\label{sec:Conclusion}

We introduced a physics-guided RL framework for determining effective sequences of real-time remedial control actions to mitigate cascading failures. The approach, focused on transmission line-switches, utilizes linear sensitivity factors to enhance RL exploration during agent training. By improving sample efficiency and yielding superior remedial control policies within a constrained computational budget, our framework ensures better utilization of grid resources, which is critical in the context of climate change adaptation and mitigation. Comparative analyses on the Grid2Op 36-bus and~{the IEEE 118-bus networks} highlight the superior performance of our framework against relevant baselines. Future work will involve using bus-split sensitivity factors~\cite{BSDF:2024} to computationally efficiently prune and identify effective bus-split actions for remedial control policy design. Another direction is to leverage the linearity of sensitivity factors to implement simultaneous remedial actions, expediting line flow control along desired trajectories.

\bibliographystyle{unsrtnat}
\bibliography{CFM}

\begin{thebibliography}{35}
\providecommand{\natexlab}[1]{#1}
\providecommand{\url}[1]{\texttt{#1}}
\expandafter\ifx\csname urlstyle\endcsname\relax
  \providecommand{\doi}[1]{doi: #1}\else
  \providecommand{\doi}{doi: \begingroup \urlstyle{rm}\Url}\fi

\bibitem[NAP(2014)]{NAPS:2004}
August $14,\;2003$ blackout: {NERC} actions to prevent and mitigate the impacts
  of future cascading blackouts.
\newblock
  \url{https://www.nerc.com/docs/docs/blackout/NERC_Final_Blackout_Report_07_13_04.pdf},
  February 2014.

\bibitem[Fisher et~al.(2008)Fisher, O'Neill, and Ferris]{Fisher:2008}
Emily~B. Fisher, Richard~P. O'Neill, and Michael~C. Ferris.
\newblock Optimal transmission switching.
\newblock \emph{IEEE Transactions on Power Systems}, 23\penalty0 (3):\penalty0
  1346--1355, 2008.

\bibitem[Khodaei and Shahidehpour(2010)]{Khodaei:2010}
Amin Khodaei and Mohammad Shahidehpour.
\newblock Transmission switching in security-constrained unit commitment.
\newblock \emph{IEEE Transactions on Power Systems}, 25\penalty0 (4):\penalty0
  1937--1945, 2010.

\bibitem[Fuller et~al.(2012)Fuller, Ramasra, and Cha]{Fuller:2012}
J.~David Fuller, Raynier Ramasra, and Amanda Cha.
\newblock Fast heuristics for transmission-line switching.
\newblock \emph{IEEE Transactions on Power Systems}, 27\penalty0 (3):\penalty0
  1377--1386, 2012.

\bibitem[Dehghanian et~al.(2015)Dehghanian, Wang, Gurrala, Moreno-Centeno, and
  Kezunovic]{Dehghanian:2015}
Payman Dehghanian, Yaping Wang, Gurunath Gurrala, Erick Moreno-Centeno, and
  Mladen Kezunovic.
\newblock Flexible implementation of power system corrective topology control.
\newblock \emph{Electric Power Systems Research}, 128:\penalty0 79--89, 2015.
\newblock ISSN 0378-7796.

\bibitem[Larsson et~al.(2002)Larsson, Hill, and Olsson]{larsson:2002}
Mats Larsson, David~J. Hill, and Gustaf Olsson.
\newblock Emergency voltage control using search and predictive control.
\newblock \emph{International Journal of Electrical Power \& Energy Systems},
  24\penalty0 (2):\penalty0 121--130, 2002.

\bibitem[Carneiro and Ferrarini(2010)]{Carneiro:2010}
Juliano S.~A. Carneiro and Luca Ferrarini.
\newblock Preventing thermal overloads in transmission circuits via model
  predictive control.
\newblock \emph{IEEE Transactions on Control Systems Technology}, 18\penalty0
  (6):\penalty0 1406--1412, 2010.

\bibitem[Almassalkhi and Hiskens(2014{\natexlab{a}})]{almassalkhi:1-2014}
Mads~R Almassalkhi and Ian~A Hiskens.
\newblock Model-predictive cascade mitigation in electric power systems with
  storage and renewables—{P}art {I}: Theory and implementation.
\newblock \emph{IEEE Transactions on Power Systems}, 30\penalty0 (1):\penalty0
  67--77, 2014{\natexlab{a}}.

\bibitem[Almassalkhi and Hiskens(2014{\natexlab{b}})]{almassalkhi:2-2014}
Mads~R Almassalkhi and Ian~A Hiskens.
\newblock Model-{p}redictive {c}ascade {m}itigation in {e}lectric {p}ower
  {s}ystems with {s}torage and {r}enewables—{P}art {II}: {C}ase-{S}tudy.
\newblock \emph{IEEE Transactions on Power Systems}, 30\penalty0 (1):\penalty0
  78--87, 2014{\natexlab{b}}.

\bibitem[Ernst et~al.(2004)Ernst, Glavic, and Wehenkel]{Ernst:2004}
D.~Ernst, M.~Glavic, and L.~Wehenkel.
\newblock Power systems stability control: reinforcement learning framework.
\newblock \emph{IEEE Transactions on Power Systems}, 19\penalty0 (1):\penalty0
  427--435, 2004.

\bibitem[Yan et~al.(2017)Yan, He, Zhong, and Tang]{Yan:2017}
Jun Yan, Haibo He, Xiangnan Zhong, and Yufei Tang.
\newblock ${Q}$-learning-based vulnerability analysis of smart grid against
  sequential topology attacks.
\newblock \emph{IEEE Transactions on Information Forensics and Security},
  12\penalty0 (1):\penalty0 200--210, 2017.

\bibitem[Duan et~al.(2020)Duan, Shi, et~al.]{Duan:2020}
Jiajun Duan, Di~Shi, et~al.
\newblock Deep-reinforcement-learning-based autonomous voltage control for
  power grid operations.
\newblock \emph{IEEE Transactions on Power Systems}, 35\penalty0 (1):\penalty0
  814--817, 2020.

\bibitem[Dwivedi and Tajer(2024)]{GRNN:2024}
Anmol Dwivedi and Ali Tajer.
\newblock {GRNN}-based real-time fault chain prediction.
\newblock \emph{IEEE Transactions on Power Systems}, 39\penalty0 (1):\penalty0
  934--946, 2024.

\bibitem[Kelly et~al.(2020)Kelly, O'Sullivan, de~Mars, and Marot]{kelly:2020}
Adrian Kelly, Aidan O'Sullivan, Patrick de~Mars, and Antoine Marot.
\newblock Reinforcement learning for electricity network operation.
\newblock \emph{arXiv:2003.07339}, 2020.

\bibitem[Marot et~al.(2020)Marot, Donnot, Romero, Donon, Lerousseau,
  Veyrin-Forrer, and Guyon]{marot:2020}
Antoine Marot, Benjamin Donnot, Camilo Romero, Balthazar Donon, Marvin
  Lerousseau, Luca Veyrin-Forrer, and Isabelle Guyon.
\newblock Learning to run a power network challenge for training topology
  controllers.
\newblock \emph{Electric Power Systems Research}, 189:\penalty0 106635, 2020.

\bibitem[Marot et~al.(2021)Marot, Donnot, Dulac-Arnold, Kelly, O’Sullivan,
  Viebahn, Awad, Guyon, Panciatici, and Romero]{marot:2021}
Antoine Marot, Benjamin Donnot, Gabriel Dulac-Arnold, Adrian Kelly, Aidan
  O’Sullivan, Jan Viebahn, Mariette Awad, Isabelle Guyon, Patrick Panciatici,
  and Camilo Romero.
\newblock Learning to run a power network challenge: {A} retrospective
  analysis.
\newblock In \emph{Proc. NeurIPS Competition and Demonstration Track}, December
  2021.

\bibitem[Rolnick et~al.(2024)Rolnick, Aspuru-Guzik, Beery, Dilkina, Donti,
  Ghassemi, Kerner, Monteleoni, Rolf, Tambe, and White]{rolnick:2024}
David Rolnick, Alan Aspuru-Guzik, Sara Beery, Bistra Dilkina, Priya~L. Donti,
  Marzyeh Ghassemi, Hannah Kerner, Claire Monteleoni, Esther Rolf, Milind
  Tambe, and Adam White.
\newblock Application-driven innovation in machine learning.
\newblock \emph{arXiv:2403.17381}, 2024.

\bibitem[Wood et~al.(2013)Wood, Wollenberg, and Shebl{\'e}]{wood:2013}
Allen~J Wood, Bruce~F Wollenberg, and Gerald~B Shebl{\'e}.
\newblock \emph{Power Generation, Operation, and Control}.
\newblock John Wiley \& Sons, 2013.

\bibitem[Donnot(2020)]{donnot:2020}
Benjamin Donnot.
\newblock {Grid2Op} - {A} {T}estbed {P}latform to {M}odel {S}equential
  {D}ecision {M}aking in {P}ower {S}ystems, 2020.
\newblock URL \url{https://github.com/rte-france/grid2op}.

\bibitem[Lan et~al.(2020)Lan, Duan, Zhang, Shi, Wang, Diao, and
  Zhang]{Lan:2020}
Tu~Lan, Jiajun Duan, Bei Zhang, Di~Shi, Zhiwei Wang, Ruisheng Diao, and Xiaohu
  Zhang.
\newblock {AI}-based autonomous line flow control via topology adjustment for
  maximizing time-series {ATC}s.
\newblock In \emph{Proc. IEEE Power and Energy Society General Meeting}, QC,
  Canada, August 2020.

\bibitem[Chauhan et~al.(2023)Chauhan, Baranwal, and Basumatary]{AAAI:2023}
Anandsingh Chauhan, Mayank Baranwal, and Ansuma Basumatary.
\newblock {PowRL}: A reinforcement learning framework for robust management of
  power networks.
\newblock In \emph{Proc. AAAI Conference on Artificial Intelligence},
  Washington, DC, June 2023.

\bibitem[Yoon et~al.(2021)Yoon, Hong, Lee, and Kim]{ICLR:2021}
Deunsol Yoon, Sunghoon Hong, Byung-Jun Lee, and Kee-Eung Kim.
\newblock Winning the {L2}{RPN} challenge: Power grid management via
  semi-{M}arkov afterstate actor-critic.
\newblock In \emph{Proc. International Conference on Learning Representations},
  May 2021.

\bibitem[Sutton et~al.(1999)Sutton, Precup, and Singh]{sutton:1999}
Richard~S Sutton, Doina Precup, and Satinder Singh.
\newblock Between {MDP}s and semi-{MDP}s: {A} framework for temporal
  abstraction in reinforcement learning.
\newblock \emph{Artificial intelligence}, 112\penalty0 (1-2):\penalty0
  181--211, 1999.

\bibitem[Ramapuram~Matavalam et~al.(2023)Ramapuram~Matavalam, Guddanti, Weng,
  and Ajjarapu]{Matavalam:2023}
Amarsagar~Reddy Ramapuram~Matavalam, Kishan~Prudhvi Guddanti, Yang Weng, and
  Venkataramana Ajjarapu.
\newblock Curriculum based reinforcement learning of grid topology controllers
  to prevent thermal cascading.
\newblock \emph{IEEE Transactions on Power Systems}, 38\penalty0 (5):\penalty0
  4206--4220, 2023.

\bibitem[Meppelink(2023)]{meppelink:2023}
Geert~Jan Meppelink.
\newblock A hybrid reinforcement learning and tree search approach for network
  topology control.
\newblock Master's thesis, NTNU, 2023.

\bibitem[Bellman(1957)]{bellman:1957}
Richard Bellman.
\newblock \emph{Dynamic Programming}.
\newblock Princeton University Press, 1957.

\bibitem[Tsitsiklis and Van~Roy(1997)]{tsitsiklis:1997}
J.N. Tsitsiklis and B.~Van~Roy.
\newblock An analysis of temporal-difference learning with function
  approximation.
\newblock \emph{IEEE Transactions on Automatic Control}, 42\penalty0
  (5):\penalty0 674--690, 1997.

\bibitem[Sutton and Barto(2018)]{sutton:2018}
Richard~S Sutton and Andrew~G Barto.
\newblock \emph{Reinforcement Learning: An Introduction}.
\newblock MIT press, 2018.

\bibitem[Mnih et~al.(2015)Mnih, Kavukcuoglu, Silver, Rusu, Veness, Bellemare,
  Graves, et~al.]{Mnih:2015}
Volodymyr Mnih, Koray Kavukcuoglu, David Silver, Andrei~A Rusu, Joel Veness,
  Marc~G Bellemare, Alex Graves, et~al.
\newblock Human-level control through deep reinforcement learning.
\newblock \emph{Nature}, 518\penalty0 (7540):\penalty0 529--533, 2015.

\bibitem[Sauer et~al.(2001)Sauer, Reinhard, and Overbye]{Sauer:2001}
P.W. Sauer, K.E. Reinhard, and T.J. Overbye.
\newblock Extended factors for linear contingency analysis.
\newblock In \emph{Proc. Hawaii International Conference on System Sciences},
  Maui, Hawaii, January 2001.

\bibitem[Quentin(2022)]{MILPAGENT:2022}
François Quentin.
\newblock {MILP}-agent, 2022.
\newblock URL \url{https://github.com/rte-france/grid2op-milp-agent}.

\bibitem[Wang et~al.(2016)Wang, , Schaul, Hessel, Hasselt, Lanctot, and
  Freitas]{wang2016dueling}
Ziyu Wang, , Tom Schaul, Matteo Hessel, Hado Hasselt, Marc Lanctot, and Nando
  Freitas.
\newblock Dueling network architectures for deep reinforcement learning.
\newblock In \emph{Proc. International Conference on Machine Learning}, New
  York, NY, June 2016.

\bibitem[Schaul et~al.(2016)Schaul, Quan, Antonoglou, and Silver]{Schaul:2016}
Tom Schaul, John Quan, Ioannis Antonoglou, and David Silver.
\newblock Prioritized experience replay.
\newblock In \emph{Proc. International Conference on Learning Representations},
  San Juan, Puerto Rico, May 2016.

\bibitem[van Dijk et~al.(2024)van Dijk, Viebahn, Cijsouw, and van
  Casteren]{BSDF:2024}
Joost van Dijk, Jan Viebahn, Bastiaan Cijsouw, and Jasper van Casteren.
\newblock Bus split distribution factors.
\newblock \emph{IEEE Transactions on Power Systems}, 39\penalty0 (3):\penalty0
  5115--5125, 2024.

\bibitem[Dwivedi et~al.(2024)Dwivedi, Paternain, and Tajer]{dwivedi:2024}
Anmol Dwivedi, Santiago Paternain, and Ali Tajer.
\newblock Blackout mitigation via physics-guided {RL}.
\newblock \emph{arXiv:2401.09640}, 2024.

\end{thebibliography}

\appendix

\section{Appendix}
\label{sec:Appendix}

\subsection{MDP Modeling}
\label{sec:MDP Modeling}
\paragraph{State Space $\mcS$:} Based on the system's state $\bX[n]$, which captures the line and bus features, we denote the MDP state at time $n$ by $\bS[n]$, defined as a moving window of the states of length $\kappa$, i.e.,
\begin{equation}
    \begin{aligned}
        \label{eq: mdp state}
        \bS[n] \dff \left[ \bX[n - (\kappa-1)], \dots, \bX[n] \right]^{\top}\ ,
    \end{aligned}
\end{equation} where the state space is $\mcS = \R^{\kappa\cdot(L\cdot N+F\cdot H)}$. Leveraging the temporal correlation of demands, decisions based on the MDP state $\bS[n]$ help predict future load demands.
\paragraph{Action Space $\mcA$:} We denote the action space by $\mcA \triangleq \mcA_{\sf line}$, where $\mcA_{\sf line}$ is the space of line-switching. Action space $\mcA_{\sf line}$ includes two actions for each line $\ell \in [L]$ associated with {reconnecting} and {removing} it. Besides these $2L$ actions, we also include a \emph{do-nothing} action to accommodate the instances at which (i)~the mandated downtime period $\tau_{\rm D}$ makes all line-switch actions operationally infeasible; or (ii)~the system's risk $\max_{i \in [L]} \rho_{i}[n]$ is sufficiently low. This action allows the agent to determine the MDP state at time $n+1$ solely based on the system dynamics driven by changes in load demand $\bD[n+1]$. 

\paragraph{Stochastic Transition Kernel $\P$:} After an action $a[n] \in \mcA$ is taken at time $n$, the  MDP state $\bS[n]$ transitions to the next state $\bS[n+1]$ according to an unknown transition probability kernel $\P$ $\bS[n+1] \sim \P(\bS \;|\;\bS[n], a[n])$ where $\P$ captures the system dynamics influenced by both the random future load demand and the implemented action $a[n]\in \mcA$.

\paragraph{Reward Dynamics $\mcR$:} To capture the immediate effectiveness of taking an action $a[n]\in \mcA$ in any given MDP state $\bS[n]$, we define an instant reward function
\begin{align}
    \label{eq: reward}
    r[n]  \dff \sum_{\ell=1}^{L} \left( 1 - \rho_{\ell}^{2}[n]  \right) - \mu_{\sf line} \left( \sum_{\ell=1}^{L} c^{\sf line}_{\ell}\cdot  W_{\ell}[n]\right)\ ,
\end{align} 
which is the decision reward associated with transitioning from MDP state $\bS[n]$ to $\bS[n+1]$, where the constant $\mu_{\sf line}$ is associated with the cost constraint $\beta_{\rm line}$ introduced in~\eqref{eq:OBJ2}, respectively. The inclusion of parameter $\mu_{\sf line}$ allows us to flexibly model different cost constraints, reflecting diverse economic considerations in power systems. Greater values for the parameter $\mu_{\sf line}$ in~\eqref{eq: reward} promote solutions that satisfy stricter cost requirements.

\subsection{Algorithmic Details}
\label{sec:Algorithmic Details}

\begin{algorithm}[h]
    \caption{Construct Set $\mcR^{\sf eff}_{\sf line}[n]$ from Action Space $\mcA^{}_{\sf line}$}
    \label{alg:algoLineRemoval}
    \begin{algorithmic}[1]
        \footnotesize
        \Procedure{Effective Set $\mcR^{\sf eff}_{\sf line}$}{$\mcA^{}_{\sf line}$} 
            \State Observe system state $\bX[n]$ and construct $\mcL[n]$
            \State Initialize $\mcR^{\sf eff}_{\sf line}[n] \leftarrow \emptyset$
            \State Construct $\mcA^{\sf rem}_{\sf line}[n] \leftarrow \{\ell \in \mcL[n]: \tau_{\rm D} = 0\;\mbox{\&}\;\tau_{\rm F} = 0\}$~\Comment{\textcolor{red}{legal removals}}
            \State Construct $\lodf[n] \in \R^{L \times L}$ matrix from $\bX[n]$ 
            \State Find $\ell_{\sf max}=\dff \argmax_{\ell \in \mcL[n]} \;\; \rho_{\ell}[n]$
            \For {line $k$ in $\mcA^{\sf rem}_{\sf line}[n]\backslash \{\ell_{\sf max}\}$}~\Comment{\textcolor{red}{legal line removals that decrease flow}}
                \State Compute $F_{\ell_{\sf max}}[n+1] \leftarrow F_{\ell_{\sf max}}[n] + \lodf_{\ell_{\sf max}, k} \cdot F_{k}[n]$ 
                    \If{$|F_{\ell_{\sf max}}[n+1]| \leq F^{\sf max}_{\ell_{\sf max}}$}
                        \State $\mcR^{\sf eff}_{\sf line}[n] \leftarrow \mcR^{\sf eff}_{\sf line}[n] \bigcup \{k\}$
                    \EndIf		
            \EndFor
            \For {line $k$ in $\mcR^{\sf eff}_{\sf line}[n]$}~\Comment{\textcolor{red}{no additional overloads}}
                \For {line $\ell$ in $\mcL[n] \backslash \{\ell_{\sf max}\}$} 
                    \State Compute $F_{\ell}[n+1] \leftarrow F_{\ell}[n] + \lodf_{\ell, k} \cdot F_{k}[n]$ 
                    \If{$|F_{\ell}[n+1]| > F^{\sf max}_{\ell}$}
                        \State $\mcR^{\sf eff}_{\sf line}[n] \leftarrow \mcR^{\sf eff}_{\sf line}[n] \backslash \{k\}$
                        \State~\textbf{Break}
                    \EndIf
                \EndFor
            \EndFor
            \State Construct $\mcA^{\sf reco}_{\sf line}[n] \leftarrow \{\ell \in \neg \mcL[n]: \tau_{\rm D} = 0\;\mbox{\&}\;\tau_{\rm F} = 0\}$~\Comment{\textcolor{red}{legal reconnect}}
            \State $\mcR^{\sf eff}_{\sf line}[n] \leftarrow \mcR^{\sf eff}_{\sf line}[n] \bigcup \mcA^{\sf reco}_{\sf line}[n]$	
            \State \Return $\mcR^{\sf eff}_{\sf line}[n]$
    \EndProcedure
    \end{algorithmic}
\end{algorithm}

\begin{algorithm}[h] 
	\caption{Physics-Guided~{Exploration}}
	\label{alg:algoExplore}
	\begin{algorithmic}[1]
            \footnotesize
		\Procedure{Physics-guided Explore}{$\mcA_{\sf line}$}
            \State Construct $\mcR^{\sf eff}_{\sf line}[n]$ from $\mcA_{\sf line}$ using Algorithm~\ref{alg:algoLineRemoval}
		\State Initialize $\maxReward \gets -\infty$
		\State Initialize $\maxAction \gets$ \textbf{None}
		\For {each action $a$ in $\mcR^{\sf eff}_{\sf line}[n]$}~\Comment{\textcolor{red}{get reward estimate}}
		\State Obtain reward estimate $\tilde r[n]$ for action $a[n]$ via sensitivity factors
		\If{$\tilde r[n] > \maxReward$}
		\State $\maxReward \gets \tilde r$
		\State $\maxAction \gets a$
		\EndIf
		\EndFor
		\State \Return $\maxAction$
		\EndProcedure
	\end{algorithmic}
\end{algorithm}

Algorithm~\ref{alg:algoLineRemoval} has three main steps. 
\begin{enumerate}
    \item The agent constructs a legal action set $\mcA^{\sf rem}_{\sf line}[n] \subset \mcA_{\sf line}$ from $\bX[n]$, comprising of permissible line removal candidates. Specifically, lines $\ell \in \mcL[n]$ with legality conditions $\tau_{\rm D} = 0$ and $\tau_{\rm F} = 0$ can only be removed rendering other control actions in $\mcA_{\sf line}$ irrelevant at time $n$. 

    \item A dynamic set $\mcR^{\sf eff}_{\sf line}[n]$ is constructed by initially identifying lines $k\in \mcA^{\sf rem}_{\sf line}[n]\backslash \{\ell_{\sf max}\}$ whose removal~\emph{decrease} flow in line $\ell_{\sf max}$ below its rated limit $F^{\sf max}_{\ell_{\sf max}}$.

    \item Finally, the agent eliminates lines from $\mcR^{\sf eff}_{\sf line}[n]$ the removal of which creates additional overloads in the network. Note that we include~\emph{all} currently disconnected lines $\ell \in \neg \mcL[n]$ as potential candidates for reconnection in the set $\mcR^{\sf eff}_{\sf line}[n]$, provided they adhere to legality conditions ($\tau_{\rm D} = 0$ and $\tau_{\rm F} = 0$). It is noteworthy that the set $\mcR^{\sf eff}_{\sf line}[n]$ is~\emph{time-varying}. Hence, depending on the current system state $\bX[n]$, $\mcR^{\sf eff}_{\sf line}[n]$ may either contain a few elements or be empty.
\end{enumerate}

\begin{algorithm}[h]
	\caption{$Q$-Guided \emph{Exploitation} with Probability $1 - \epsilon_{n}$}
	\label{alg:algoExploit}
	\begin{algorithmic}[1]
            \footnotesize
		\Procedure{$Q$-guided Exploit}{$\mcA_{\sf line}, \btheta_{n}$}
            \State Infer MDP state $\bS[n]$ from $\bX[n]$
            \State Construct $\mcA^{\sf legal}_{\sf line}[n] \leftarrow \{\ell \in [L]: \tau_{\rm D} = 0\;\mbox{\&}\;\tau_{\rm F} = 0\}$~\Comment{\textcolor{red}{legal line-switch}}
		\State $\mcA^{\sf legal} \leftarrow \mcA^{\sf legal}_{\sf line}[n] \bigcup \mcA_{\sf gen}$
  		\State Initialize $\bQ[n] \leftarrow$ DQN$_{\btheta_{n}}(\bS[n])$ 
		\State $\bQ_{\mcA^{\sf legal}}[n] \leftarrow \text{Filter}(\bQ[n], \mcA^{\sf legal})$\Comment{\textcolor{red}{filter \emph{legal} $Q$-values}}
		\State $\topFiveActions \leftarrow \text{TopFive}(\bQ_{\mcA^{\sf legal}}[n])$\Comment{\textcolor{red}{find top-$5$ legal $Q$-values}}
		\State Initialize $\maxReward \gets -\infty$
		\State Initialize $\maxAction \gets$ \textbf{None}
		\For {each action $a$ in $\topFiveActions$}~\Comment{\textcolor{red}{get reward estimate}}
		      \State Obtain reward estimate $\tilde r[n]$ for action $a$ via flow model~\eqref{eq: lodf} 
			\If{$\tilde r[n] > \maxReward$}
			\State $\maxReward \gets \tilde r[n]$
			\State $\maxAction \gets a$
			\EndIf
		\EndFor
		\State \Return $\maxAction$
		\EndProcedure
	\end{algorithmic}
\end{algorithm}

\paragraph{\bf $Q$-Guided Exploitation Policy (Algorithm~\ref{alg:algoExploit})} 
The agent refines its action choices over time by leveraging the feature representation $\btheta_{n}$, learned through the minimization of the temporal difference error via stochastic gradient descent. Specifically, the agent employs the current DQN$_{\btheta_{n}}$ to select an action~$a\in \mcA$ with probability $1 - \epsilon_{n}$. The process begins with the agent inferring the MDP state $\bS[n]$ in~\eqref{eq: mdp state} from $\bX[n]$. Next, the agent predicts a $\bQ[n]\in \R^{|\mcA|}$ vector using the network model DQN$_{\btheta_{n}}(\bS[n])$ through a forward pass, where each element represents $Q$-value predictions associated with each remedial control actions $a[n]\in \mcA$. Rather than choosing the action with the highest $Q$-value, the agent first identifies legal action subset $\mcA^{\sf legal} \dff \mcA^{\sf legal}_{\sf line}[n]$ from $\bX[n]$. Next, the agent identifies actions $a[n]\in \mcA^{\sf legal}$ associated with the top-$5$ $Q$-values within this legal action subset $\mcA^{\sf legal}$ and chooses one optimizing for the reward estimate $\tilde r[n]$. This policy 
accelerates learning without the need to design a sophisticated reward function $\mcR$ that penalizes illegal actions.

\subsection{Grid2Op Environment Details}
\label{sec:Grid2Op Environment Details}

\paragraph{Grid2Op Environment} Grid2Op is an open-source gym-like platform for simulating power transmission networks with real-world operational constraints. Grid2Op offers diverse episodes throughout the year with distinct monthly load profiles. Each episode encompasses generation $\bG[n]$ and load demand $\bD[n]$ set-points for all time steps $n\in[T]$ across every month throughout the year. Each episode represents approximately 28 days with a 5-minute time resolution, based on which we have horizon $T=8062$. December consistently shows high aggregate demand, pushing transmission lines closer to their maximum flow limits while May experiences relatively lower demand. 

\paragraph{Datasets} For both systems, we have performed a~\emph{random} split of Grid2Op episodes. For the test sets, we selected 32 scenarios for the Grid2Op 36-bus system and 34 scenarios for the IEEE 118-bus system, while assigning 450 scenarios to the training sets and a subset for validation to determine the hyperparameters. To ensure proper representation of various demand profiles, the test set includes at least two episodes from each month.

\paragraph{Performance Metrics:} A key performance metric is the agent's survival time ST$(T)$, averaged across all test set episodes for $T=8062$. We explore factors influencing ST through analyzing action diversity and track~\emph{unique} control actions per episode. Furthermore, we quantify the fraction of times each of the following three possible actions are taken: ``do-nothing," and ``line-switch $\mcA_{\sf line}$,". {Since the agent takes remedial actions only under~\emph{critical} states associated with critical time instances $n$, we report action decision fractions that exclusively stem from these critical states, corresponding to instances when $\rho_{\ell_{\sf max}}[n] \geq \eta$. We also note that monthly load demand variations $\bD[n]$ influence how frequently different MDP states $\bS[n]$ are visited. This results in varying control actions per episode. To form an overall insight, we report the \emph{average} percentage of actions chosen across all test episodes.

\begin{table}[h]
	\centering
	\scalebox{0.9}{
		\begin{tabular}{|c | c | c | c|} 
			\hline
			System-State Feature $\bX[n]$ & Size & Type & Notation \\ [0.4ex] 
			\hline\hline
			\texttt{prod\_p} & $G$ & float & $\bG[n]$ \\ 
			\hline
			\texttt{load\_p} & $D$ & float & $\bD[n]$ \\ 
			\hline
			\texttt{p\_or, p\_ex} & $L$ & float & $F_{\ell}[n]$ \\ 
			\hline
			\texttt{a\_or, a\_ex} & $L$ & float & $A_{\ell}[n]$ \\ 
			\hline
			\texttt{rho} & $L$ & float & $\rho_{\ell}[n]$ \\ 
			\hline
			\texttt{line\_status} & $L$ & bool & $\mcL[n]$ \\ 
			\hline
			\texttt{timestep\_overflow} & $L$ & int & overload time \\ 
			\hline
			\texttt{time\_before\_cooldown\_line} & $L$ & int & line downtime \\ 
			\hline
			\texttt{time\_before\_cooldown\_sub} & $N$ & int & bus downtime \\ 
			\hline
		\end{tabular}
	}
	\caption{Heterogeneous input system state features $\bX[n]$.}
	\label{tab:inputFeatures}
\end{table}

\paragraph{System Parameters and MDP State Space} The Grid2Op 36-bus system consists of $N=36$ buses, $L=59$ transmission lines (including transformers), $G=10$~\emph{dispatchable} generators, and  $D=37$ loads. We employ $F=8$ line and $H=3$ bus features (Table~\ref{tab:inputFeatures}), totaling $O =567$~\emph{heterogeneous} input system state $\bX[n]$ features. Each MDP state $\bS[n]$ considers the past $\kappa = 6$ system states for decision-making. Without loss of generality, we set $\eta=0.95$ specified in Section~\ref{sec: Problem Formulation} as the threshold for determining whether the system is critical. 

The IEEE 118-bus system consists of $N=118$ buses, $L=186$ transmission lines, (including transformers), $G=32$~\emph{dispatchable} generators, and $D=99$ loads. While in principle we can choose all the 11 features in Table~\ref{tab:inputFeatures}, to improve the computational complexity associated with agent training, we choose a subset of line-related features, specifically, $F=5$ line features ($\texttt{p\_or, a\_or, rho, line\_status}$ and $\texttt{timestep\_overflow}$). This results in a total of $O = 930$~\emph{heterogeneous} input system state features and consider the past $\kappa = 5$ system states for decision-making. Without loss of generality, we set $\eta=1.0$.

After performing a line-switch action $a[n]\in\mcA_{\sf line}$ on any line $\ell$, we impose a mandatory downtime of $\tau_{\rm D}=3$ time steps (15-minute interval) for each line $\ell \in [L]$. In the event of natural failure caused due to an overload cascade, we extend the downtime to $\tau_{\rm F}=12$ (60-minute interval).

\paragraph{MDP Action Space - Line-Switch Action Space Design $\mcA_{\sf line}$:} Following the MDP modeling discussed in Section~\ref{sec:MDP Modeling}, for the Grid2Op 36-bus system we have $|\mcA_{\sf line}|=119\;(2L + 1)$ and {for the IEEE 118-bus system we have  $|\mcA_{\sf line}|=373\;(2L + 1)$.}

\subsection{Baseline Agents}
\label{sec:Baseline Agents}

For the chosen performance metrics, we consider~{four} alternative baselines: (i)~$\noop$ agent consistently opts for the ``do-nothing" action across all scenarios, independent of the system-state $\bX[n]$; (ii) $\reco$ agent decides to ``re-connect" a disconnected line that greedily~\emph{maximizes} the reward estimate $\tilde r[n]$~\eqref{eq: reward} at the current time step $n$. In cases where reconnection is infeasible due to line downtime constraints or when no lines are available for reconnection, the $\reco$ agent defaults to the ``do-nothing" action for that step;~{(iii)~\texttt{milp\_agent}\cite{MILPAGENT:2022} agent strategically minimizes over-thermal line margins using line switching actions $\mcA_{\rm line}$ by formulating the problem as a mixed-integer linear program (MILP); and (iv)~RL + Random Explore baseline agent: we employ a DQN$_{\btheta}$ network with a~\emph{tailored} random $\epsilon_{n}$-greedy exploration policy during agent~{training}. Specifically, similar to Algorithm~\ref{alg:algoExplore}, the agent first constructs a legal action set $\mcA^{\sf legal}_{\sf line}[n] \dff \{\ell \in [L]: \tau_{\rm D} = 0 , \tau_{\rm F} = 0\}$ from $\bX[n]$ at critical times. In contrast to Algorithm~\ref{alg:algoExplore}, however, this agent chooses a~\emph{random} legal action in the set $a[n] \in \mcA^{\sf legal}_{\sf line}[n]$ (instead of using $\mcR^{\sf eff}_{\sf line}[n]$). In the Grid2Op 36-bus system, using this random exploration policy, we train the DQN$_{\btheta}$ for $20$ hours of repeated interactions with the Grid2Op simulator for each $\mu_{\sf line} \in \{0, 0.5, 1, 1.5\}$. We report results associated with the~\emph{best} model $\btheta$ and refer to the best policy obtained following this random $\epsilon_{n}$-greedy exploration by $\pi^{\sf rand}_{\btheta}(\mu_{\sf line})$. Similarly, in the IEEE 118-bus system, we train the DQN$_{\btheta}$ model for 15 hours of repeated interactions.

\subsection{DQN Architecture and Training}
\label{sec:DQN Architecture and Training}

Our DQN architecture features a feed-forward NN with two hidden layers, each having $O$ units and adopting~\texttt{tanh} nonlinearities. The input layer, with a shape of $|\bS[n]| = O \cdot \kappa$, feeds into the first hidden layer of $O$ units, followed by another hidden layer of $O$ units. The network then splits into two streams: an advantage-stream $\bA_{\btheta}(\bS[n], \cdot)\in \R^{|\mcA|}$ with a layer of $|\mcA|$ action-size units and~\texttt{tanh} non-linearity, and a value-stream $V_{\btheta}(\bS[n]) \in \R$ predicting the value function for the current MDP state $\bS[n]$. $\bQ_{\btheta}(\bS[n], \cdot)$ are obtained by adding the value and advantage streams. We penalize the reward function $r[n]$ in~\eqref{eq: reward} in the event of failures attributed to overloading cascades and premature scenario termination ($n < T$). Additionally, we normalize the reward constraining its values to the interval $[-1, 1]$. For the Grid2Op 36-bus system, we use a learning rate $\alpha_{n}=5\cdot 10^{-4}$ decayed every $2^{10}$ training iterations, a mini-batch size of $B=64$, an initial $\epsilon = 0.99$ exponentially decayed to $\epsilon = 0.05$ over $26\cdot 10^{3}$ agent-MDP training interaction steps and choose $\gamma = 0.99$. {Likewise, for the IEEE 118-bus system we use similar parameters with a mini-batch size of $B=32$.} {Likewise, for the IEEE 118-bus system we set $\alpha_{n}=9\cdot 10^{-4}$ with a mini-batch size of $B=32$ and $21\cdot 10^{3}$ agent MDP training interaction steps.}

\subsection{Results for the IEEE 118-bus System}
\label{sec:118}

\begin{table*}[h]
	\centering
	\renewcommand{\arraystretch}{1.3} 
	\scalebox{0.75}{
		{\begin{tabular}{|c|c|c|c|c|c|c|c|}
				\hline
				\thead{Action Space $(|\mcA|)$}  
				& \thead{Agent Type}
				& \thead{Avg. ST} 
				& \thead{$\%$\\ Do-nothing}
				& \thead{$\%$\\ Reconnect}
				& \thead{$\%$\\ Removals}
				& \thead{Avg. Action\\ Diversity}\\
				\hline
				\hline
				$-$                      &$\noop$ &$4371.91$ &$100$    &$-$     &$-$     &$-$ \\
				\hline
				$\mcA_{\sf line}\;(187)$ &$\reco$ &$2813.64$ &$98.73$  &$1.26$    &$-$   &$1.235\;(0.66\%)$ \\
				\hline
				$\mcA_{\sf line}\;(373)$ &\texttt{milp\_agent}\cite{MILPAGENT:2022}  &$4003.85$ &$15.64$ &$0.88$  &$83.46$      &$5.617\;(1.505\%)$ \\
				\hline
				\multirow{2}{*}{\rotatebox[origin=c]{0}{\makecell{$\mcA_{\sf line}\;(373)$}}}
				&\multirow{1}{*}{\rotatebox[origin=c]{0}{\makecell{RL\;+ Random Explore}}} 
				&$4812.88$  &$3.58$ &$20.30$ &$76.08$  &$8.323\;(2.231\%)$\\
				\cline{2-7}
				&\multirow{1}{*}{\rotatebox[origin=c]{0}{\makecell{RL\;+ Physics Guided Explore}}} 
				&$\mathbf{5767.14}$  &$1.86$ &$25.34$ &$72.77$  &$\mathbf{16.235\;(4.352\%)}$ \\
				\hline
		\end{tabular}}
	}
	\caption{Performance on the IEEE 118-bus system with $\eta = 1.0$ and $\mu_{\sf line}=0$.}
	\label{tab:lineAndGen_118}
\end{table*}

All the results for the IEEE 118-bus system are tabulated in Table~\ref{tab:lineAndGen_118}. Starting from the baselines, we observe that the $\noop$ agent achieves a significantly higher average ST of 4,371 steps, compared to the $\reco$ agent's 2813.64 steps. This observation highlights the importance of strategically selecting~\emph{look-ahead} decisions, particularly in more complex and larger networks. Contrary to common assumptions, the $\reco$ agent's greedy approach of reconnecting lines can instead reduce ST, demonstrating that $\noop$ can be more effective.

Focusing on the line switch action space $\mcA_{\sf line}$, we observe that the agent with policy $\pi^{\sf rand}_{\btheta}$ survives 4812.88 steps, a $10.1\%$ increase over baselines, by allocating $76.08\%$ to remedial control actions for line removals. More importantly, our physics-guided policy $\pi^{\sf physics}_{\btheta}$ achieves an average ST of 5767 steps, a $31.9\%$ increase over baselines and a $19.2\%$ improvement compared to $\pi^{\sf rand}_{\btheta}$ with greater action diversity. Fig.~\ref{fig:36bus_interactions} illustrates the number of agent-MDP interactions as a function of training time, showcasing that the physics-guided exploration is more thorough for a given computational budget.

While this paper focuses on the improvements achieved through effective exploration using action space $\mcA_{\sf line}$, further enhancements of the physics-guided design can be realized by extending the action space to generator adjustments, i.e., $\mcA_{\sf line} \cup \mcA_{\sf gen}$. As presented in the study~\cite{dwivedi:2024}, this extension allows for a richer exploration of the state space. It enables reaching additional states by taking actions $a[n] \in \mcA_{\sf gen}$ from states that were originally accessible only via actions $a[n] \in \mcA_{\sf line}$, thereby improving downstream performance.

\begin{figure}[h]
    \centering
    \includegraphics[width=\linewidth]{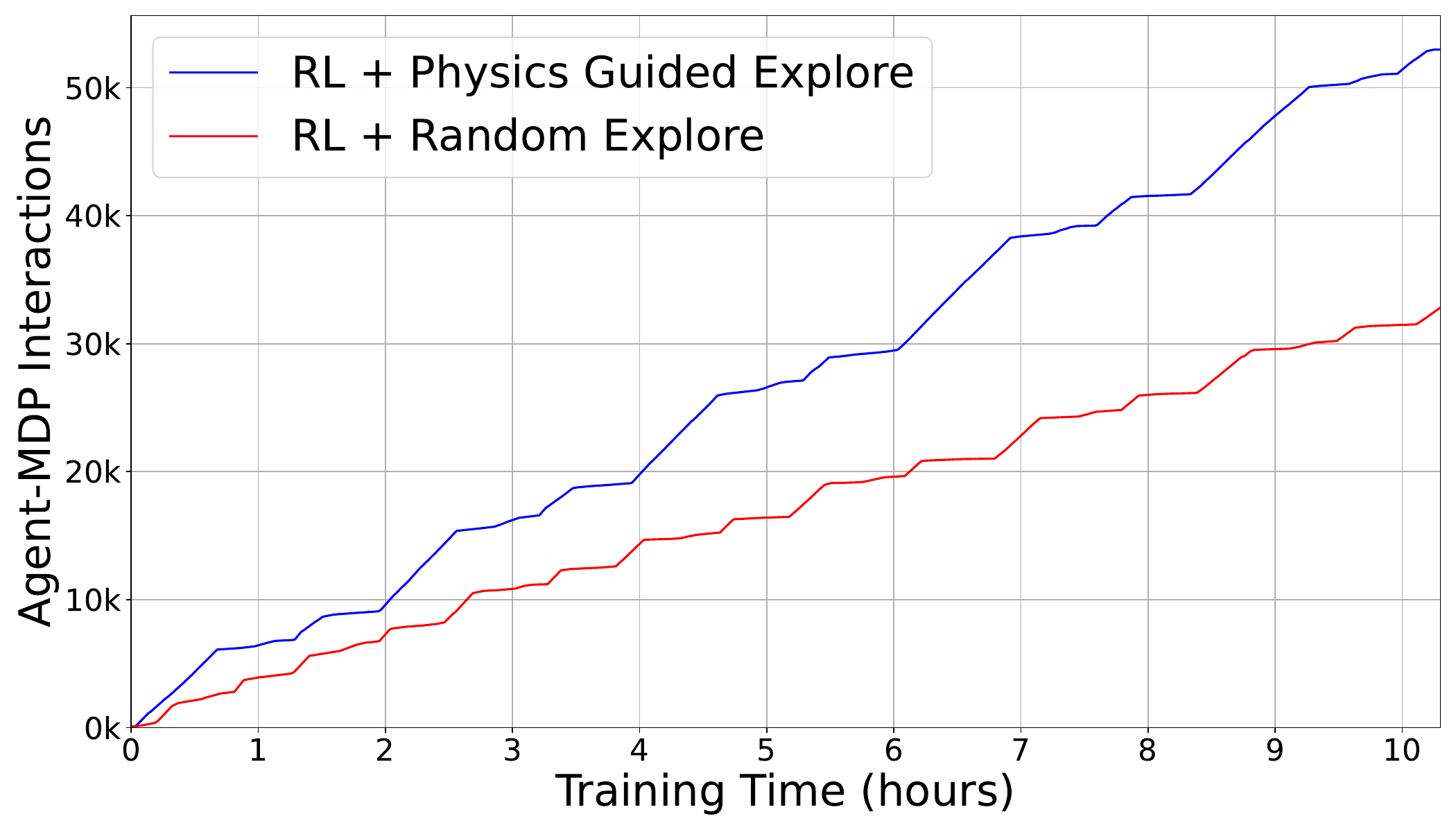}
    \caption{Agent$-$MDP interactions for the IEEE 118-bus system with $\eta = 1.0$ and $\mu_{\sf line}=0$.}
    \label{fig:36bus_interactions}  
\end{figure}




\end{document}